\documentclass[letterpaper, 10 pt, conference]{ieeeconf}
\IEEEoverridecommandlockouts    

\usepackage{multirow}
\usepackage{amsmath}
\usepackage{amssymb}
\usepackage[ruled,vlined]{algorithm2e}
\usepackage{graphicx}
\usepackage{tikz}
\graphicspath{{./Figures/}}
\usepackage[pagebackref,breaklinks]{hyperref}
\usepackage{xspace}
\usepackage{booktabs}
\usepackage{subcaption}
\usepackage{url}

\newcommand{\dataset}{VZCrash\@\xspace}  

\newcommand{\circled}[1]{%
  \begin{tikzpicture}[baseline=(char.base)]
    \node[shape=circle, draw=none, fill=gray, text=white, inner sep=1pt, minimum size=1.3em, font=\footnotesize\bfseries] (char) {\footnotesize #1};
  \end{tikzpicture}%
}

\title{\dataset: A Large-Scale IMU Dataset of Ego-Vehicle Crashes}

\author{
 	\parbox{\textwidth}{%
 		\centering
 		Tommaso Bianconcini$^{1}$, Henrique Piñeiro Monteagudo$^{1}$, Aurel Pjetri$^{1}$, Tomaso Trinci$^{1}$, Leonardo Taccari$^{1}$%
 	}%
 	\thanks{$^{1}$Verizon Connect, Florence, Italy. 
        }
    \thanks{
 		Email: {\tt tommaso.bianconcini@verizonconnect.com}%
        }
    \thanks{\copyright 2026 IEEE. Personal use of this material is permitted. Permission from IEEE must be obtained for all other uses, in any current or future media.}
    }

\begin{document}
	
	\maketitle
	\thispagestyle{empty}
	\pagestyle{empty}
	
	\begin{abstract}

We introduce \dataset, the largest publicly available dataset of real-world vehicle collision data featuring 
Inertial Measurement Unit (IMU) telemetry. The dataset contains more than 31,000 validated crashes and 158,000 negative samples, including hard cases and distractors.
Each sample includes acceleration and angular velocity at 100 Hz, and GPS speed at 1 Hz.
Events in \dataset were captured by devices installed on a fleet of 73,010 commercial vehicles of different sizes driving in the United States over the span of several years.

We also present an extensive experimental study enabled by the volume of the dataset. 
We first benchmark several different approaches, from a simple threshold-based heuristic to state-of-the-art deep learning models.
Then, we present an experiment demonstrating the importance of scaling data to train high-quality crash detection models, and we show that scale is especially important when these models need to be deployed into a real-world environment.
\dataset is publicly available at this URL: \url{https://huggingface.co/datasets/vzc-research-chapter/VZCrash}.
	\end{abstract}
	
	\section{Introduction}
	\label{sec:introduction}

The advancement of intelligent transportation systems for crash detection has traditionally relied on data to bridge the gap between lab tests and the complexities of real-world environments.
With the aim of improving road safety, the research community has developed several large-scale datasets.
However, a significant gap remains between visual datasets and those providing the inertial telemetry necessary to understand the physics of an impact.

Several existing datasets have attempted to address vehicle safety and  analysis, yet they remain limited in scale or modalities. The SHRP2 dataset \cite{hankey2016description} contains only 1,541 verified crashes, is not publicly available, and features a low IMU sampling rate of 10~Hz. 
Autonomous driving datasets such as nuScenes~\cite{caesar2020nuscenes} or BDD100k~\cite{yu2020bdd100k} provide large-scale multi-modal data, but do not contain actual vehicle crashes or safety-critical events. 
Other recent benchmarks, such as the Nexar collision prediction dataset~\cite{moura2025nexar} and CAP-DATA \cite{fang2022cognitive}, focus on visual anticipation of accidents but lack the high-frequency IMU telemetry essential for physical impact analysis. 
The US-Accident dataset~\cite{moosavi2019accident} is a very large-scale database, containing 2.25 million cases of traffic crashes that occurred within the United States between 2016 and 2019. However, this dataset provides only high-level geographical features, 
that can be used for statistical analysis or identification of high-risk areas, but not for real-time detection.

To address these limitations, we introduce \dataset, the largest publicly available ego-centric dataset of vehicle dynamics with real-world collisions. Collected from a footprint of 73,010 commercial vehicles across the United States over several years (from 2020 to 2025), \dataset comprises almost 190,000 unique events, including a corpus of above 31,000 verified crashes. 

\begin{figure}[!t]
    \centering
    \includegraphics[width=\linewidth, trim={0 0 0 4cm}, clip]{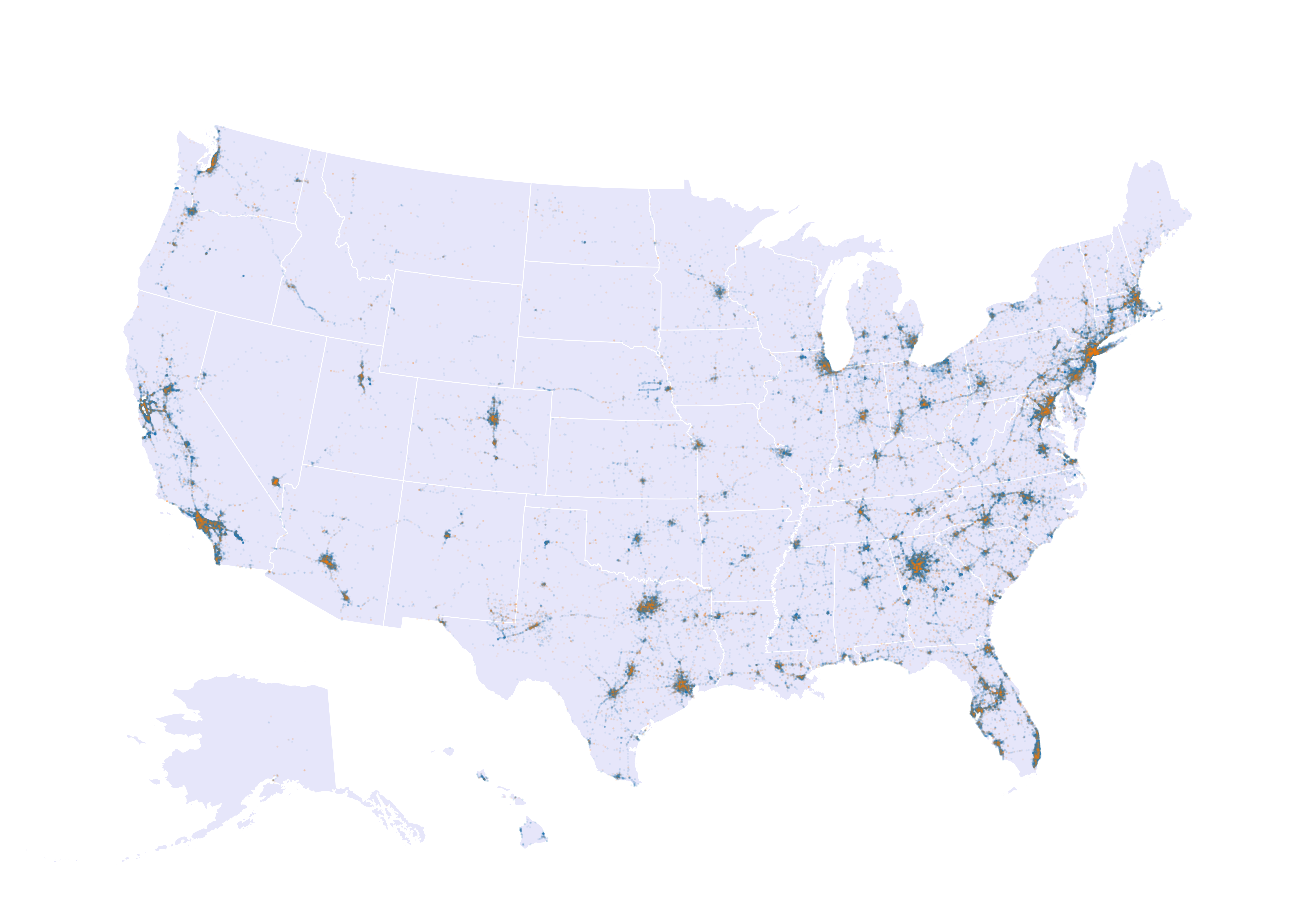}
    \caption{Geographical distribution of \dataset events. Crashes in orange.}
    \label{fig:geo_distribution}
\end{figure}

Scale is essential because vehicle collisions are rare and follow a long-tail distribution.
Considering that, as per the Fatality Analysis Reporting System from NHTSA~\cite{nhtsa_fars_2023}, the rate in the US is around 117 per 100 million km, we can estimate that a fleet of 100,000 vehicles driving 100~km per day could generate around 10 crashes per day.
Hence, capturing them requires sustained long-term observation and curation. \dataset provides the necessary volume to capture the variance of real-world crashes, including difficult to observe edge cases.  
Unlike curated or synthetic datasets, our collection also captures the raw ``noise" of connected devices \emph{in the wild}, including sensor miscalibration, signal bias, and high-frequency vibrations from varied mounting conditions, road surfaces, and vehicle types. 

A second central contribution of this work is an extensive experimental study enabled by our dataset.
We benchmark several baselines and deep learning approaches on the crash detection task. 
Then, we study the effect of scaling training data on the performance of machine learning models for this task. Our results demonstrate that the ability of models to generalize across diverse real-world conditions scales significantly with data volume. 
We show that increasing the supervised training set to the full dataset leads to substantial improvements in model robustness, suggesting that crash detection from acceleration data benefits from this magnitude of data, previously unavailable in the public domain.

    \section{\dataset dataset}
	\label{sec:dataset}

\dataset contains almost 190,000 samples, including 31,090 verified crashes. For each event, we release 16 seconds of 100~Hz tri-axial accelerometer and gyroscope data captured by dashcam-integrated inertial measurement units and 1~Hz GPS speeds.
IMUs periodically calibrate themselves based on the direction of movement of the vehicles, so that the $X$-axis points forward, the $Y$-axis points leftward and the $Z$-axis points upward and measures 1~g at rest. Recordings were triggered on-device based on multiple threshold heuristics on the acceleration signals: high values on the $X$-axis are considered hard braking and hard acceleration events, high $XY$-norm values are considered shock events. All these events were further processed on the cloud using an ensemble of specialized machine learning models (cf.~\cite{kubin2021deep,taccari2018}). The events classified as positives were reviewed along with a random subsample of the negatives.
The events were collected across the United States between 2020 and 2025 as shown in Fig.~\ref{fig:geo_distribution}.

Each event in \dataset has been reviewed by expert human reviewers, trained for this particular task, with access to video footage from the dashcam in addition to the IMU data. The final labels are the result of a 3-way consensus.
We define a \textit{crash} as an event involving the ego-vehicle that satisfies at least one of the following criteria:
\begin{itemize}
    \item \textbf{Unintentional contact}: Any physical impact between the ego-vehicle and external entities, such as other vehicles, animals, or infrastructure (e.g., traffic barriers), independently from the fault attribution.
    \item \textbf{Accidental departure}: Unintentional departure from the established roadway or drivable surface.
\end{itemize}
We distinguish crashes from other high-intensity acceleration events that exhibit similar kinematic profiles. Such ``kinematic distractors'' include pothole encounters, deliberate off-road maneuvers, and commercial trailer docking or coupling procedures. We explicitly exclude these from the crash category and label them as \emph{negative} examples. 
Another specific class of negative examples that might contain acceleration profiles with strong vibrations or decelerations are the so-called \textit{near-miss}.
Following \cite{hankey2016description}, we define a \textit{near-miss} as any traffic situation requiring a rapid evasive maneuver by the ego-vehicle to successfully avoid a crash.

\begin{figure}[!ht]
    \centering
    \includegraphics[width=\linewidth]{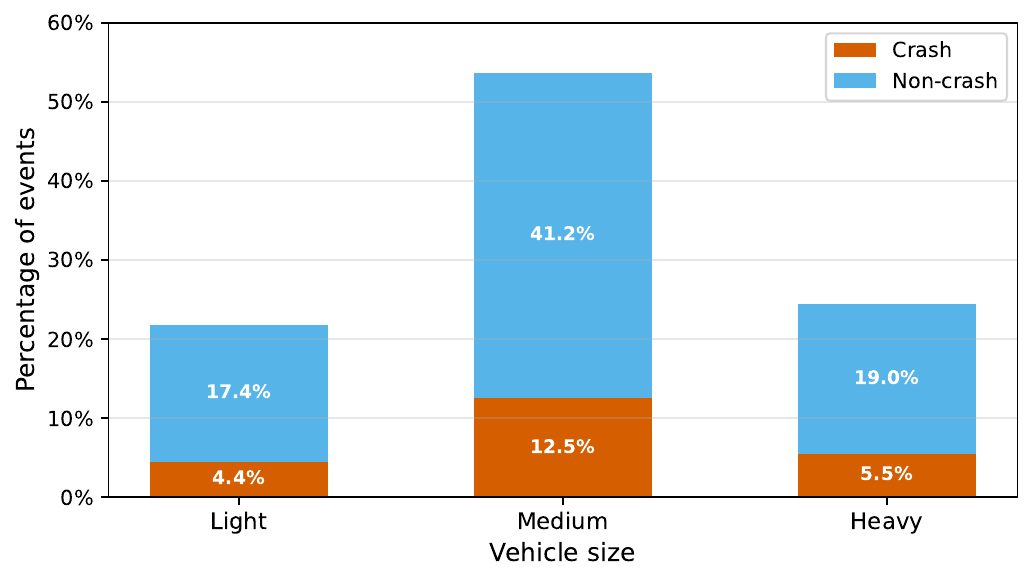}
    \caption{Event distribution by vehicle size. Data reflects the subset (approx. 70\%) for which vehicle size is available.}
    \label{fig:vehicle_size_distribution}
\end{figure}

Unlike synthetic or smaller datasets, \dataset captures the ``noise'' of the real world. It spans 73,010 vehicles and therefore reflects a high degree of spatial and mechanical diversity, including:
\begin{itemize}
    \item \textit{vehicle heterogeneity} - a wide range of vehicle sizes, from cars and light delivery vans to heavy-duty trucks, see Fig.~\ref{fig:vehicle_size_distribution};
    \item \textit{installation variance} - diverse dashcam mounting positions and orientations;
    \item \textit{environmental noise} - real-world artifacts such as sensor miscalibration, high-frequency vibrations caused by degraded mounts or poor road conditions.
\end{itemize}

\begin{figure*}[t]
    \centering
    \includegraphics[width=\linewidth]{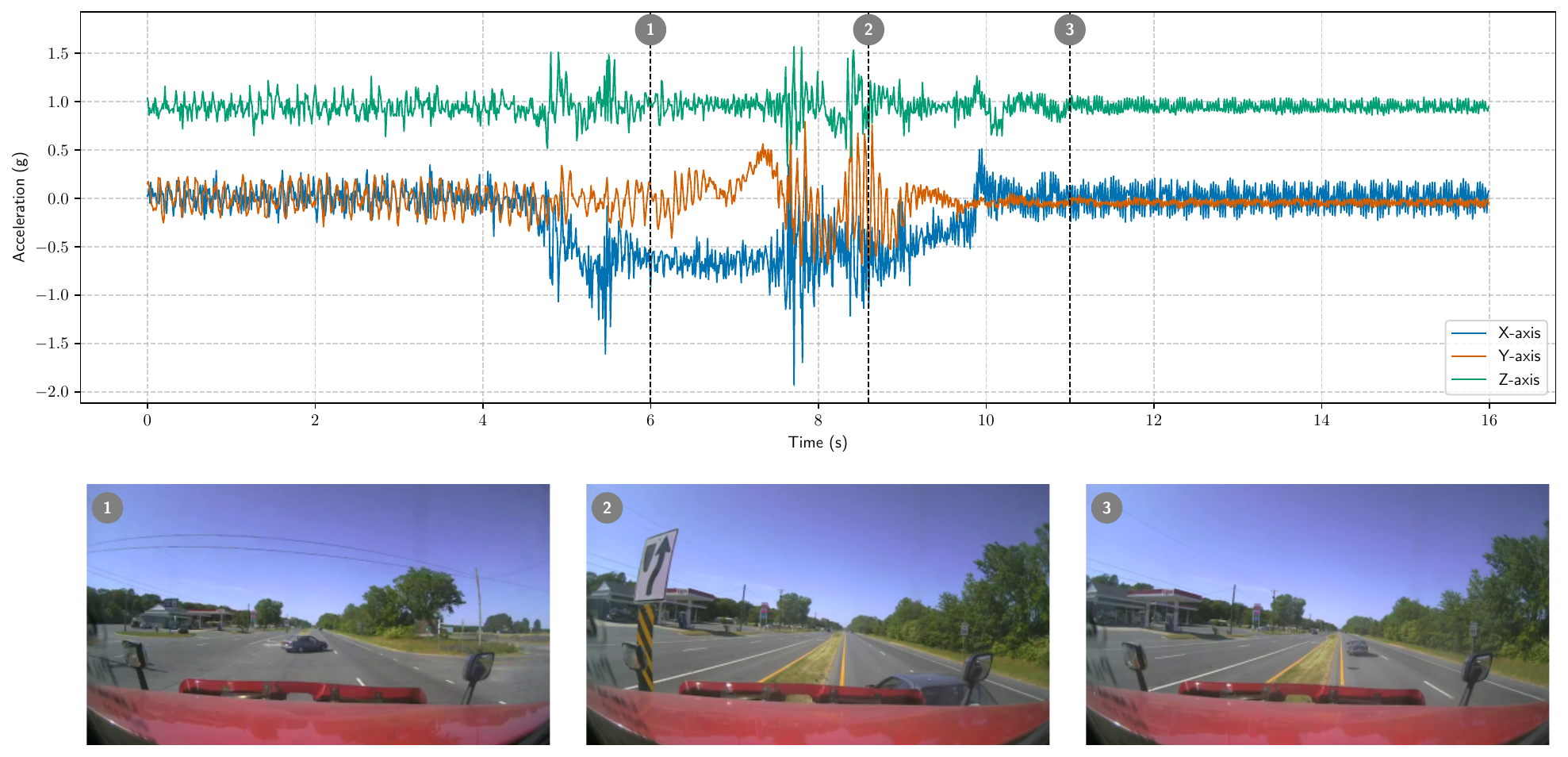}
    \caption{
    Example of a collision in \dataset. An oncoming vehicle enters the ego-vehicle’s lane from a perpendicular road. The driver applies the brakes \protect\circled{1}, but the collision remains unavoidable \protect\circled{2}. The ego-vehicle subsequently comes to a rest in the median strip \protect\circled{3}.
    }
    \label{fig:crash_example}
\end{figure*}

Figure~\ref{fig:crash_example} illustrates a crash from \dataset together with corresponding video frames. This vehicle is heavy-duty and, as can be seen from the acceleration plot, the signal while the vehicle is driving is quite noisy. Once the vehicle stops, the vibrations, especially on the lateral axis, greatly decrease.

\begin{figure*}[t]
    \centering
    \begin{subfigure}[b]{0.32\textwidth}
        \centering
        \includegraphics[width=\textwidth]{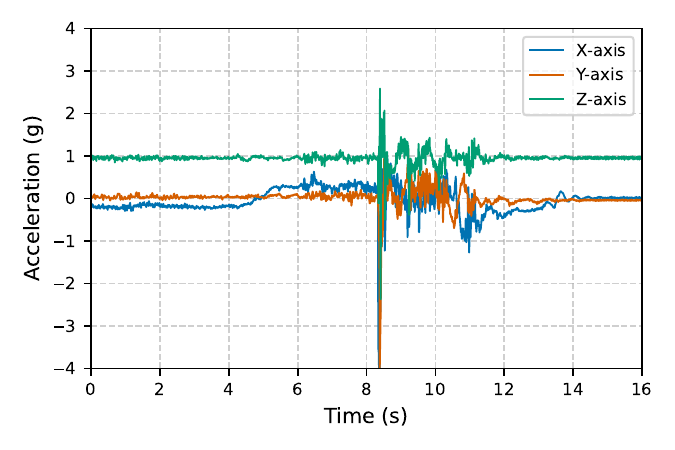}  
        \caption{Frontal Impact}
        \label{fig:front_impact}
    \end{subfigure}
    \hfill
    \begin{subfigure}[b]{0.32\textwidth}
        \centering
        \includegraphics[width=\textwidth]{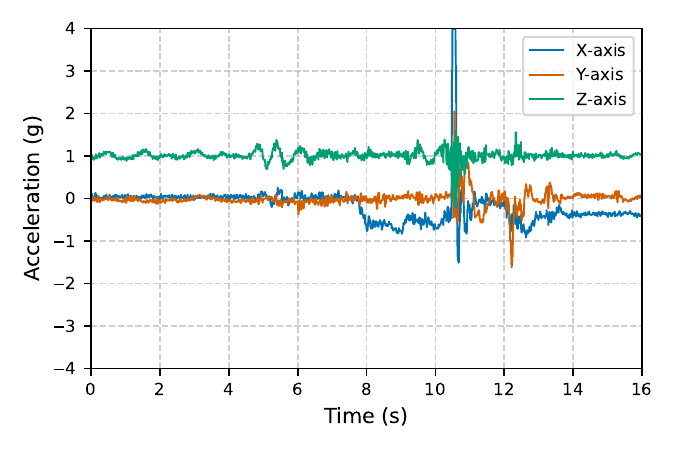}  
        \caption{Rear-end Collision}
        \label{fig:rear_impact}
    \end{subfigure}
    \hfill
    \begin{subfigure}[b]{0.32\textwidth}
        \centering
        \includegraphics[width=\textwidth]{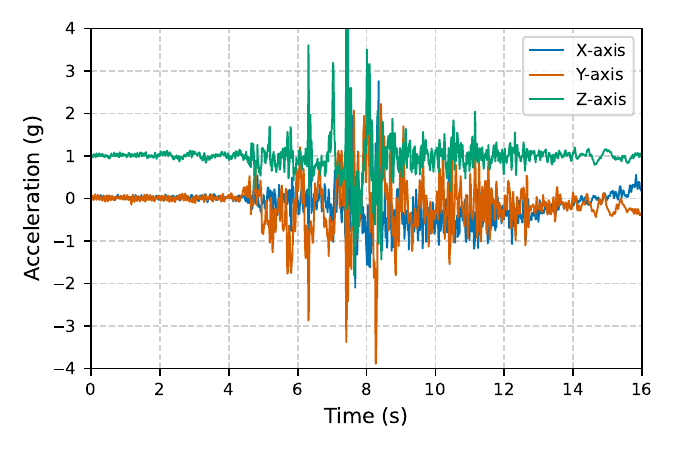}  
        \caption{Roadway Departure}
        \label{fig:road_departure}
    \end{subfigure}

    \vspace{1em} 

    \begin{subfigure}[b]{0.32\textwidth}
        \centering
        \includegraphics[width=\textwidth]{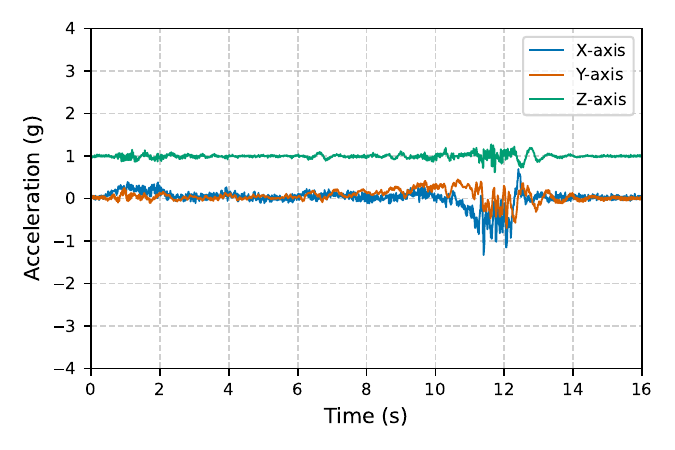}  
        \caption{Near-Miss (Hard Braking)}
        \label{fig:near_miss}
    \end{subfigure}
    \hfill
    \begin{subfigure}[b]{0.32\textwidth}
        \centering
        \includegraphics[width=\textwidth]{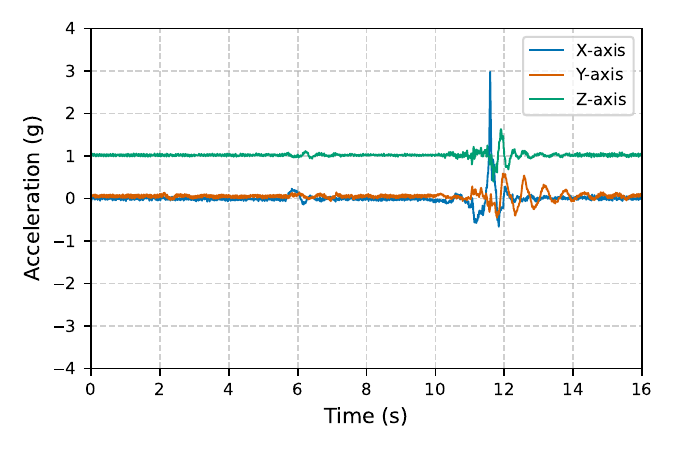}  
        \caption{Distractor: Truck Docking}
        \label{fig:trailer_hitch}
    \end{subfigure}
    \hfill
    \begin{subfigure}[b]{0.32\textwidth}
        \centering
        \includegraphics[width=\textwidth]{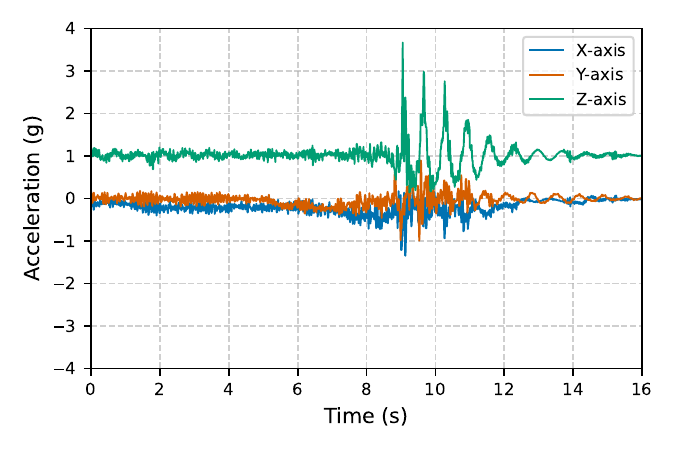}  
        \caption{Distractor: Irregular Road Surface}
        \label{fig:pothole}
    \end{subfigure}

    \caption{Representative tri-axial accelerometer traces from \dataset. The top row illustrates confirmed crash events characterized by high-magnitude impulses or sustained loss of control. The bottom row highlights the difficulty of the task, showing a near-miss and two common ``kinematic distractors'' (trailer docking and potholes) that exhibit high-g signatures but do not constitute a crash.}
    \label{fig:accel_examples}
\end{figure*}

Figure~\ref{fig:accel_examples} presents accelerometer data for a number of interesting events. All of them contain some kind of accelerometer spike, showing that not all \emph{negative} examples are trivial to classify.

	\section{Experimental Study}
	\label{sec:experiments}

\subsection{Methods}
\label{sec:sota_results}

We conduct a series of training experiments employing a range of different architectures and models, either designed for crash detection or adapted from related domains. While \dataset includes gyroscope and GPS speed telemetry to support future multimodal research, our benchmarking focuses exclusively on accelerometer data. A vehicle collision is fundamentally defined by a sudden, high-magnitude transfer of kinetic energy, making tri-axial acceleration the most critical and direct signal for impact detection.


\subsubsection*{Physical baseline} We first establish a simple physical baseline, where the score is defined as the peak across time of the vector norm of the acceleration along the longitudinal and transversal vehicle axes ($X$ and $Y$ in our data). 
This is a learning-free approach inspired by what is commonly done both in GPS tracking devices to detect ``impulses", and algorithms that control airbag deployment.

\subsubsection*{CNN-RNN}
We adapt the architecture proposed for the task of crash classification in~\cite{kubin2021deep}, which utilizes a convolutional encoder to extract high-level spatial features from raw IMU signals. These features are subsequently processed by a GRU Recurrent Neural Network (RNN)~\cite{gru} to capture the temporal vehicle dynamics necessary for crash classification.

\subsubsection*{1D Swin Transformer}
This model is adapted from the two-stream DUST framework~\cite{shi2024dust}. 
The original architecture employs a dual-branch approach using a Video Swin Transformer~\cite{swin} and a 1D Swin Transformer to fuse visual and telemetry data.
We isolate and retain the 1D Swin Transformer branch to evaluate its effectiveness in processing purely inertial streams.

\subsubsection*{CNN-Transformer}
Taking inspiration from the hierarchical feature extraction of the previous models, we design a hybrid CNN-Transformer architecture. In this setup, a convolutional branch serves as a learnable tokenizer for the inertial signal. The resulting tokens are processed by two layers of multi-head self-attention. We employ a class token to aggregate the global temporal context, which is then forwarded to a dense layer for final classification.

\subsubsection*{Chronos-Bolt}
We leverage a foundational model approach by fine-tuning the encoder of Chronos-Bolt~\cite{ansari2024chronos}, originally designed for time-series forecasting. We pair the encoder with a classification head composed of two fully connected layers. To ensure the physical magnitude of the collision signal is preserved, we modify the original architecture by replacing Chronos's standard instance normalization with a dataset-wide standardization based on global mean and standard deviation.

\subsubsection*{Scalogram Classifier}
Taking inspiration from what is commonly done in audio pattern recognition~\cite{kong2020panns,gong2021ast} and in human activity recognition from IMU sensors~\cite{luo2025har,sakai2025driver}, we encode the three accelerometer streams as scalograms, a time-frequency representation of the signal obtained by taking the magnitude of their continuous wavelet transform (CWT). 
The three scalograms are then fed as RGB images to an image classification model. 
In these experiments we use MobileNetV3 Small~\cite{howard2019searching} for its compact size and efficiency. 
Given the significant domain shift from the classical image classification task, we train the model from scratch (random initialization).


\subsection{Benchmark on the Full \dataset Dataset}

The dataset is partitioned into training (72\%), validation (14\%), and test (14\%) sets.
To ensure statistical robustness, all models are trained three times on the full training split. We use an early stopping policy, monitoring the Average Precision (AP), also known as the Area Under the Precision Recall Curve, on the validation set to prevent overfitting.
Training is performed on Nvidia T4 GPUs with 16GB memory. 

\begin{table}[hb]
\centering
\caption{Benchmark results on the \dataset test set of models trained on the full training set. AP scores are reported as the mean across three independent training runs. Best in \textbf{bold} and second best \underline{underlined}.}
\label{tab:model_recap}
\begin{tabular}{@{}lrrr@{}}
\toprule
\textbf{Algorithm} & \textbf{Parameters}  & \textbf{Latency (ms)}  &\textbf{AP (\%)} \\
\midrule
Physical Baseline & --- & \textbf{0.5} & 89.62 \\
CNN-RNN & 1.7 M & 29.4 & \textbf{97.56} \\
1D Swin Transformer & 189.7 M & 215.6& 91.76 \\
CNN-Transformer & 1.3 M & \underline{7.2}& \underline{97.54} \\
Chronos-Bolt & 8.8 M & 28.3& 96.64 \\
Scalogram Classifier & 2.5 M & 57.5& 96.23  \\ \bottomrule
\end{tabular}
\end{table}

An initial analysis of the results in Table~\ref{tab:model_recap} reveals that the physical baseline achieves an AP of nearly 90\%. 
This high performance suggests that the task is fairly simple to solve for a large fraction of cases, and indeed threshold-based algorithms are still used in many safety applications that require low power and low latency, such as insurance telematics or Automated Crash Notification (ACN) systems. 
While the effectiveness of this simple kinematic heuristic is evident, the performance gap between the baseline and the evaluated deep learning architectures is significant. The best performing models appear to be the CNN-RNN architecture and the hybrid CNN-Transformer one, both reaching around 97.5\% of AP with a small number of parameters.

It is worth commenting on the poor performance of the larger-scale model, 1D Swin Transformer. Architectures based on transformers are known to lack the inductive biases of CNN, and we posit that this model might need a larger-scale dataset or a self-supervised pre-training phase to be competitive with the other ones we tested. We consider further investigation out of the scope of this work.

Table~\ref{tab:model_recap} also reports measurements of the time it takes to process a 16-second raw acceleration signal on a single CPU core of a server equipped with an Intel Xeon P-8259CL @2.5~GHz. In addition to the model inference latency, this includes signal pre-processing steps, such as filtering, normalization, and, for the Scalogram Classifier, the extraction of the spectrograms.\footnote{The implementation of all the methods is in python with numpy, scipy, and pytorch, and could be further optimized.}
The baseline is obviously the fastest; all other algorithms, with the exception of 1D Swin Transformer, are fairly low latency and could be deployed in most modern embedded devices. Among the others, the CNN-Transformer is the most efficient architecture, outperforming the other models by a factor of $3\times$. Conversely, the Scalogram Classifier is the slowest, largely due to its extensive preprocessing requirements.

\begin{figure}
    \centering
    \includegraphics[width=\linewidth]{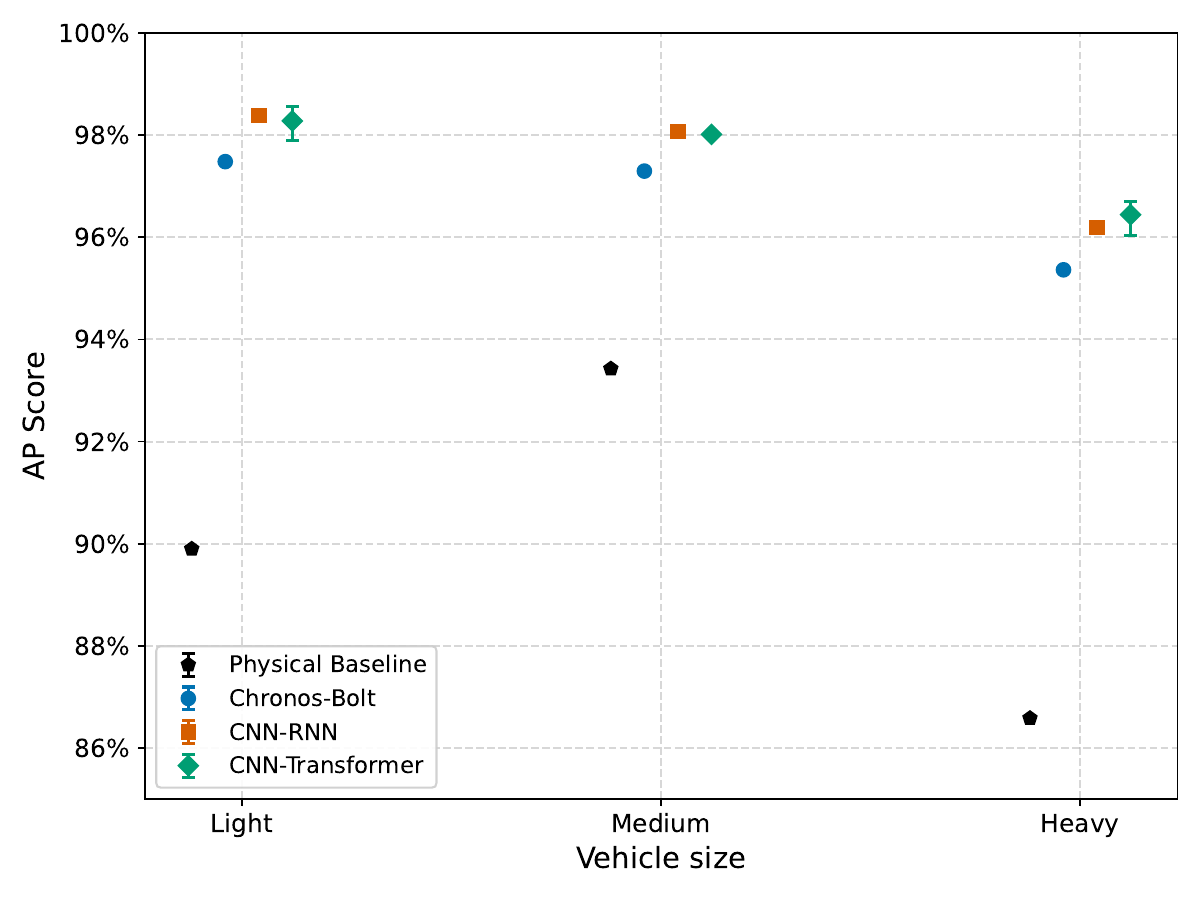}
    \caption{AP score per vehicle size cohort.}
    \label{fig:aps_by_vehiclesize}
\end{figure}

Figure~\ref{fig:aps_by_vehiclesize} compares the performance of the top three models and the physical baseline across different vehicle size cohorts in the test set. AP decreases for the largest vehicles due to increased signal noise and their greater structural mass, which attenuates the acceleration profile of low-energy collisions.

\subsection{Model Sensitivity to Data Volume}

The results in Table~\ref{tab:model_recap} 
demonstrate that all evaluated architectures achieve good levels of performance when trained on the full supervised dataset.  
However, in order to collect a sufficiently large number of meaningful examples, it is necessary to have a huge and varied pool of vehicles that drive for a significant amount of time, and to afford the operational overhead of collecting, reviewing and annotating hundreds of thousands of events, if not millions. 

Given the high cost of this process, we seek to understand the marginal utility provided by collecting more data, and to quantify the performance-to-volume relationship. 
We design an experiment in which we simulate scenarios of data scarcity by training on progressively smaller subsets of the \dataset training partition, and testing on the full test set. 
Specifically, we evaluate model performance using 1\%, 5\%, 10\%, 25\%, and 50\% of the available supervised samples.

To have a fair and rigorous comparison, we keep the same class ratio, subsampling equally both negative and positive examples; the same subsampled partitions are used for all model architectures and across all independent training runs; and we maintain the same hyperparameter configurations, optimization strategies, and early stopping criteria used in the full-scale experiments. This analysis allows us to determine the minimum data threshold required to distinguish true collisions from kinematic distractors and to assess whether high-capacity architectures require the full scale of \dataset to reach their theoretical performance ceiling.

\begin{figure}
    \centering
    \includegraphics[width=\linewidth]{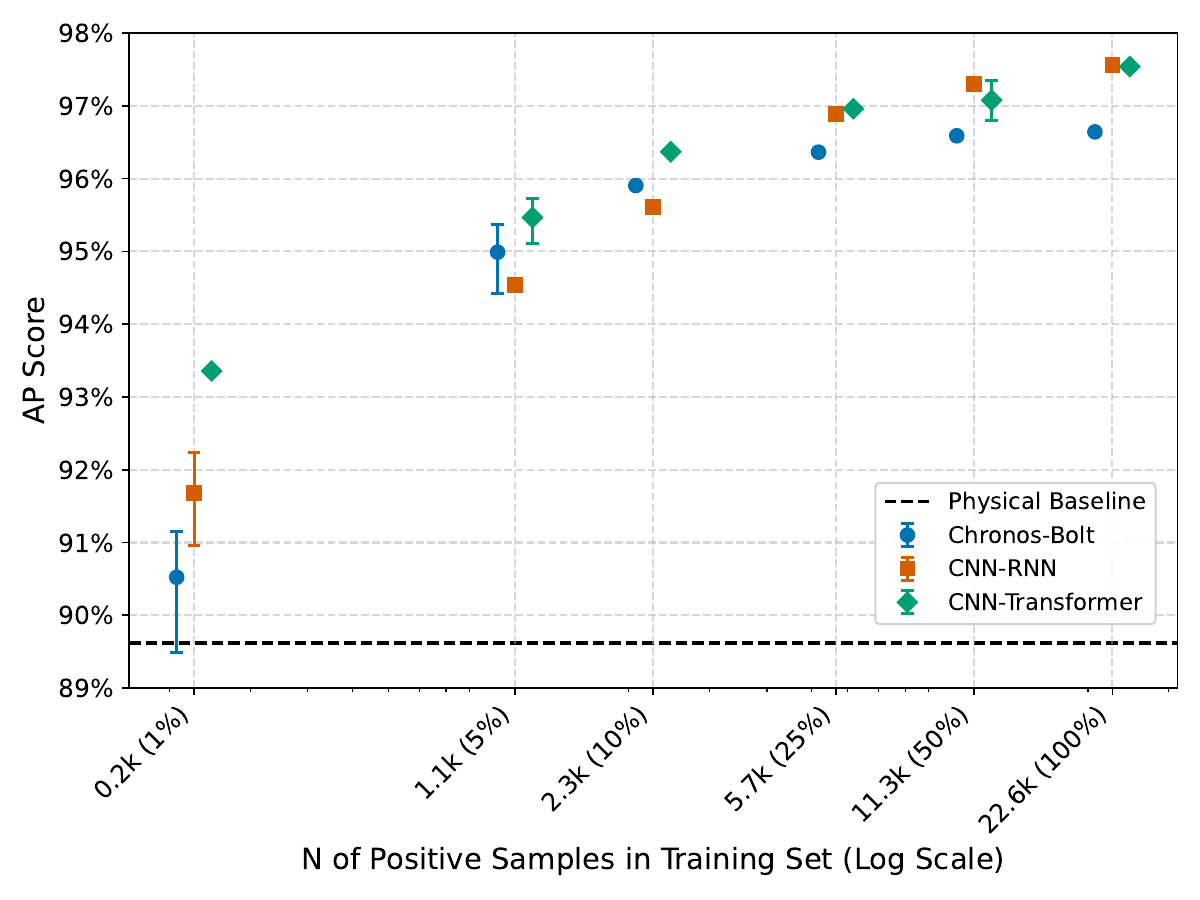}
    \caption{
    AP on the \dataset test set of the best three models obtained with progressively larger training sets.
    The $x$ axis is in logarithmic scale.
    }
    \label{fig:aps_at_percentages}
\end{figure}

Figure~\ref{fig:aps_at_percentages} shows the AP on the test set of the 3 best performing models, Chronos-Bolt, CNN-RNN, and CNN-Transformer, trained with progressively larger training sets, compared to the physical baseline as a reference.
We can see that using only 1\% of the training set, that corresponds to only around 200 collisions and 800 negative examples, the performance of the ML approaches is already better than the baseline, especially the CNN-Transformer which, being the lighter architecture, seems to achieve better performances with respect to the other models with the lowest data volumes.
A clear improvement appears once we include at least 1,000 positive examples. 
The marginal utility of new examples decreases as we get closer to the full size, but both CNN-RNN and CNN-Transformer have still not fully plateaued, showing that further gains might be obtained with an even larger dataset.



\subsection{Benchmark on a Real-World Event Population}
\label{sec:real_world_ditribution}

The \dataset dataset contains a fairly balanced class ratio (roughly 84-16\%).
In real-world applications, collision events are characterized by their extreme rarity. 
Releasing a dataset with a realistic distribution would be infeasible: we would need to release tens of millions of negative events.

We conduct an experiment where we estimate the performance of models trained on \dataset under realistic deployment conditions, with a long-tail distribution of events.
We collected Harsh Driving Events (HDEs) from a fleet of approximately 123,000 vehicles over a 48-hour period in February 2026, resulting in a corpus of 735,000 events. 

Upon manual verification, only 143 events are confirmed as actual crashes -- less than 0.02\% of the total volume. To identify these rare positives efficiently, we employ an ensemble-informed labeling strategy: we review the 1,000 events that received the highest confidence scores across all models previously trained on \dataset.


\begin{figure}
    \centering
    \includegraphics[width=\linewidth]{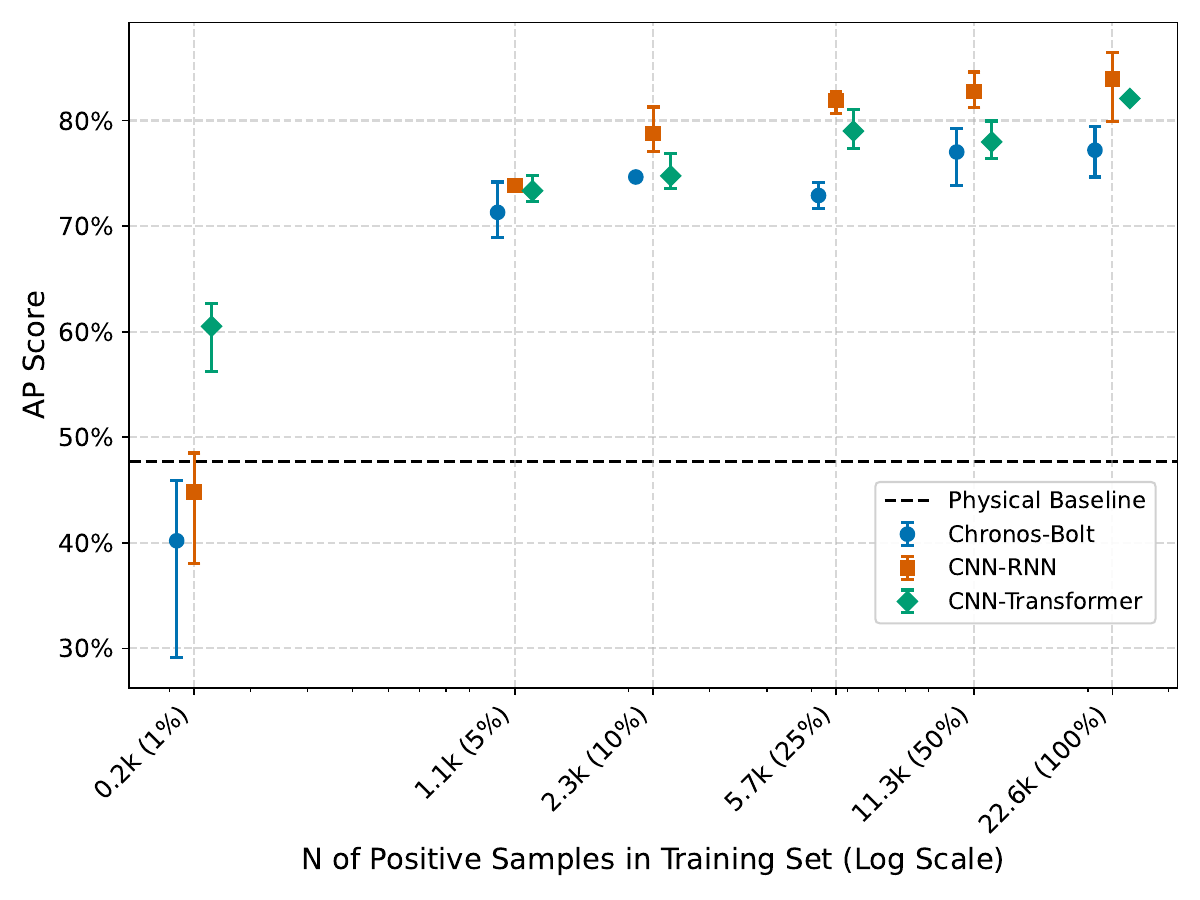}
    \caption{AP on the 735k-event real-world population of the best three models trained on \dataset with progressively larger training sets. 
    The $x$ axis is in logarithmic scale.}
    \label{fig:aps_real_distribution}
\end{figure}

In Fig.~\ref{fig:aps_real_distribution} we show that the Average Precision of the 
best model drops significantly, from 97.5\% on the \dataset test set, to around 86\% \emph{in the wild}.
Interestingly, models trained on a smaller amount of data suffer even more, and for Chronos-Bolt and CNN-RNN the AP drops below 50\%, worse than the simple physical baseline.
CNN-Transformer proves better in a low-data regime.


\begin{figure}
    \centering
    \includegraphics[width=\linewidth]{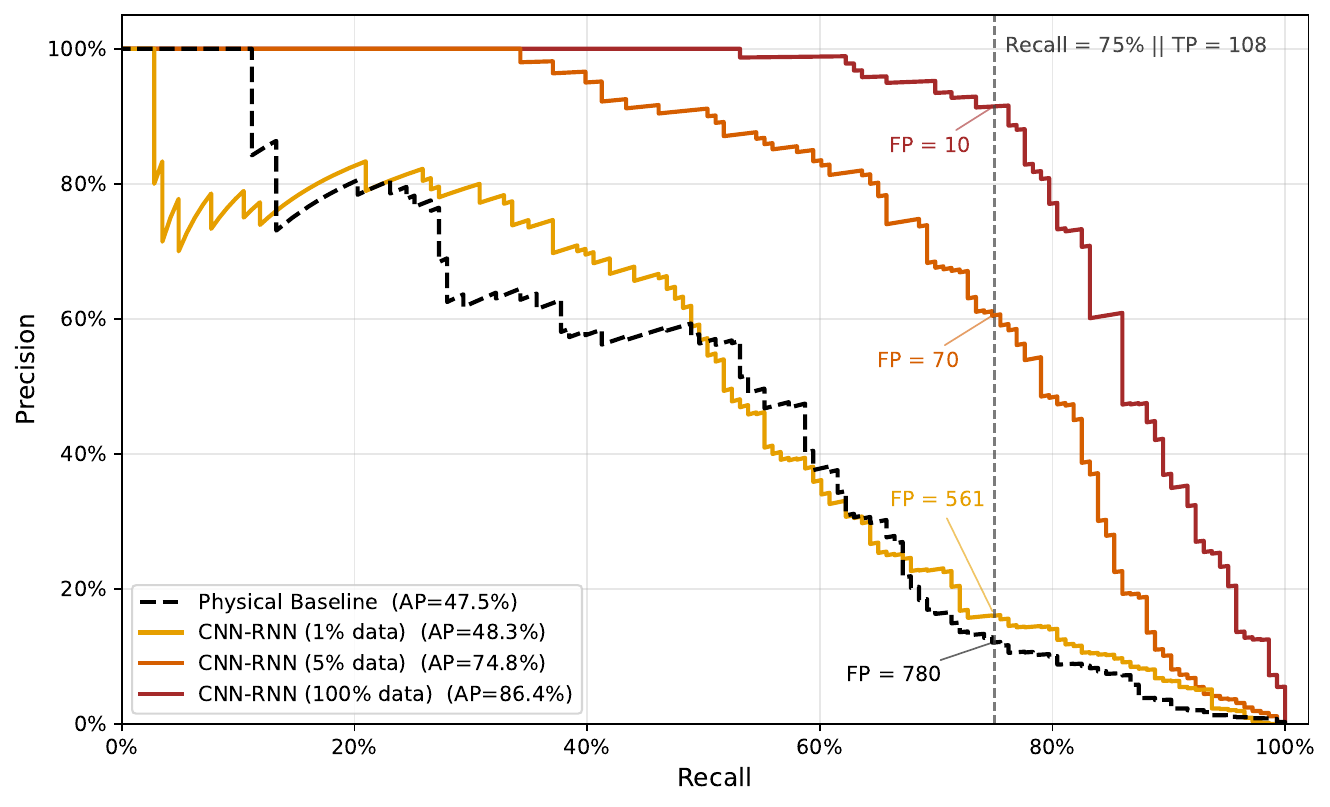}
    \caption{Precision-Recall curves on the 735k-event real-world population of CNN-RNN models trained on \dataset with different training set sizes and the Physical Baseline.}
    \label{fig:pr_at_08_recall_two_day}
\end{figure}

Figure~\ref{fig:pr_at_08_recall_two_day} further illustrates the impact of data volume on model performance. Consider a scenario where a 75\% recall target is required (detecting 108 out of 143 crashes on the 735k-event real-world population). At this threshold, the Physical Baseline generates 780 false positives, resulting in a low precision of 12\%. Performance remains poor for the CNN-RNN trained on only 1\% of the dataset, which still yields 561 false positives. However, precision improves substantially when the training set is increased to 5\% (70 false positives), and reaches 91.5\% (only 10 false positives) when using the full training set. 

This experiment underscores that the task of crash detection is much more difficult than it appears at first glance: 
a model must maintain near-perfect precision to avoid detecting a huge amount of false positives in a stream where non-crash events outnumber actual collisions by a factor of 5,100 to 1.


	\section{Summary and Conclusions}
	\label{sec:conclusion}
We introduce \dataset, the largest publicly available dataset of real-world vehicle collisions featuring high-frequency IMU telemetry, and conduct extensive numerical experiments that underline the importance of large and diverse data to effectively address the problem of crash detection.
In addition to accelerometer data at 100Hz, \dataset provides synchronized gyroscope readings and 1Hz GPS-derived speed. Integrating angular velocity and speed data could enhance detection accuracy, particularly in complex edge cases where accelerometer data alone may be ambiguous.

While the current version of the dataset focuses on kinematic data, there is clear potential for expansion. 
Adding frames or video data would provide useful context for these events, though releasing these data remains challenging due to the significant associated privacy risks.
Beyond additional data modalities, future updates to \dataset will focus on increasing label granularity. Specifically, we plan to move from binary crash detection to labeling specific event types, such as the direction of impact or the severity of the collision.

We hope that the release of \dataset as a public benchmark will support research in collision detection. By providing an alternative to private or simulated datasets, our aim is to facilitate the development of models that are better equipped to handle the noise and extreme class imbalance inherent in real-world telemetry.

	
	\bibliographystyle{IEEEtran}
	\bibliography{root} 
	
\end{document}